\def\year{2019}\relax
\documentclass[letterpaper]{article} %DO NOT CHANGE THIS
\usepackage{aaai19}  %Required
\usepackage{times}  %Required
\usepackage{helvet}  %Required
\usepackage{courier}  %Required
\usepackage{url}  %Required
\usepackage{graphicx}  %Required
\usepackage{algorithm}
\usepackage{algorithmic}
\usepackage{amsmath}
\usepackage{bm}
\usepackage{multirow}
\usepackage{stfloats}
\usepackage{caption,subcaption}
\newcommand{\D}{\mathcal{D}}

\begin{document}
% The file aaai.sty is the style file for AAAI Press 
% proceedings, working notes, and technical reports.
%
\title{Every Node Counts: Self-Ensembling Graph Convolutional Networks \\ for Semi-Supervised Learning}

\author{Yawei Luo$^{1,2}$,\hspace{2mm} Tao Guan$^{1}$,\hspace{2mm} Junqing Yu$^1$,\hspace{2mm} Ping Liu$^{2,3}$,\hspace{2mm} Yi Yang$^2$ \vspace{0.5cm} \\
$^1$School of Computer Science \& Technology, Huazhong University of Science \& Technology\\
$^2$CAI, University of Technology Sydney\\
$^3$JD.COM Silicon Valley Research Center, Big Data Group
% For a paper whose authors are all at the same institution,
% omit the following lines up until the closing ``}''.
% Additional authors and addresses can be added with ``\and'',
% just like the second author.
% To save space, use either the email address or home page, not both
% \and
% Second Author\\
% Institution2\\
% First line of institution2 address\\
% {\tt\small secondauthor@i2.org}
}

\maketitle
\begin{abstract}
Graph convolutional network (GCN) provides a powerful means for graph-based semi-supervised tasks. However, as a localized first-order approximation of spectral graph convolution, the classic GCN can not take full advantage of unlabeled data, especially when the unlabeled node is far from labeled ones. To capitalize on the information from unlabeled nodes to boost the training for GCN, we propose a novel framework named Self-Ensembling GCN (SEGCN), which marries GCN with Mean Teacher -- another powerful model in semi-supervised learning. SEGCN contains a student model and a teacher model. As a student, it not only learns to correctly classify the labeled nodes, but also tries to be consistent with the teacher on unlabeled nodes in more challenging situations, such as a high dropout rate and graph collapse. As a teacher, it averages the student model weights and generates more accurate predictions to lead the student. In such a mutual-promoting process, both labeled and unlabeled samples can be fully utilized for backpropagating effective gradients to train GCN. In three article classification tasks, \emph{i.e.} Citeseer, Cora and Pubmed, we validate that the proposed method matches the state of the arts in the classification accuracy.
%By combining Mean Teacher and GCN, we improve the baseline method on Cora with 20 labels each class from 81.5\% to 83.1\%, and on Citeseer with 20 labels each class from 70.3\% to 72.4\%, which achieves a new state of the art in terms of node classification accuracy.
\end{abstract}

\section{Introduction} \label{sec:introduction}
\noindent Semi-supervised learning (SSL) aims to build a better classifier, by utilizing huge amounts of unlabeled data which is readily accessible, together with a limited number of labeled data. Such line of work is of great significance because it achieves a high accuracy while requiring less human effort for data annotation. Recently, SSL has gained considerable attention when applied to deep learning-based methods, which are well known for their high demand on sizable and reliable labeled samples. Through distilling knowledge from unlabeled data, SSL boosts the deep learning-based methods to a new level in many tasks, \emph{e.g.,} speech recognition~\cite{dai2015semi}, image segmentation~\cite{papandreou2015weakly} and video understanding~\cite{caelles2017one}.

Inspired by the great success of SSL on regular Euclidean-based data such as speech, images, or video, a surge of recent approaches seek to apply SSL to data in a more general form -- graph. The motivation is natural: in many real problems, the data samples are on irregular grid or more generally in non-Euclidean domains, e.g. point cloud~\cite{wang2018dynamic}, chemical molecules~\cite{li2018adaptive}  and social networks~\cite{rahimi2018semi}. Instead of regularly shaped tensors, those data are better to be structured as graph, which is capable of handling varying neighborhood vertex connectivity. Similar to the original goal on regular data, SSL on graph-structured data aims to classify the all the nodes in a graph using a small subset of labeled nodes and large amounts of unlabeled nodes. The recently developed graph convolutional neural network 
\begin{figure}
\centering
\includegraphics[width=0.90\linewidth]{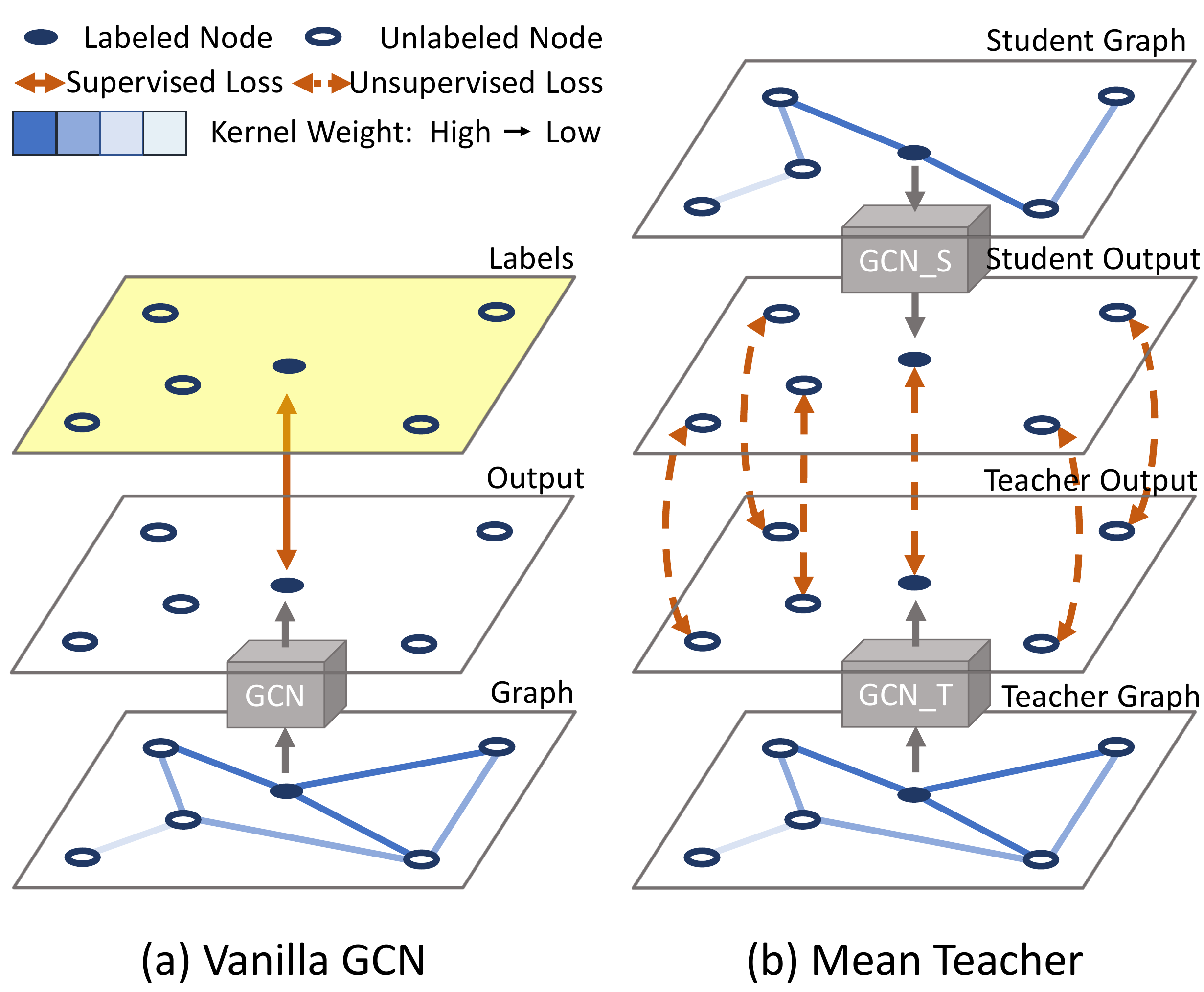}
\caption{(Best viewed in color.) \textbf{(a)} Vanilla GCN, where the each node gathers features from its neighbors in limited scope and only labeled nodes are supervised under the classification loss. \textbf{(b)} Mean Teacher model, where student model operates under a perturbed setting and tries to be consistent with the predictions of teacher model.\\
Our method marries \textbf{(a)} and \textbf{(b)}, which utilizes both supervised classification loss and unsupervised consistency loss to train GCN. Such framework enables us to explore more unlabeled knowledge to boost the classification accuracy under the semi-supervised setting.}
\end{figure}
(GCNN)~\cite{defferrard2016GCNN} and the following graph convolutional network (GCN)~\cite{kipf2016semi} are successful attempts along this line, which generalize the powerful convolutional neural network (CNN) in Euclidean data to modeling graph-structured data. This line of work capitalizes on the relation between nodes and enables the features to propagate between neighboring vertices. During the training, the supervised loss acts upon the confluent features of labeled node and then distribute the gradient information across other unlabeled adjacent nodes. Such mechanism makes the features of both labeled and unlabeled vertices in the same cluster similar, thus largely easing the classification task.  

Although it has made great progresses, GCN fails to take full advantage of unlabeled data, especially when the unlabeled node is far away from labeled ones. This is due to a $K$-layer GCN only captures node information up to $K$-hop neighborhood, which cannot effectively propagate the information to the entire graph. Taking the two-layers GCN as an example. In such network, a vertex $v_i$ would \emph{directly} aggregate and filter features in its $2$-hop neighborhood $N_{v_i}$. Considering that an adjacent vertex $v_j \in N_{v_i}$ also aggregates features from its own $2$-hop neighborhood $N_{v_j}$, the $v_i$ can \emph{indirectly} discover farther vertices beyond $N_{v_i}$. However, such indirect link is \emph{de facto} negligible which is represented with a tiny kernel weight. Consequently, any unlabeled vertex with shortest path distance $> 2$ from labeled ones in graph would gain very limited attention and are prone to be underutilized during the training. A deeper network with more graph convolutional layers may help to discover information in such remote nodes. However, as mentioned in~\cite{li2018deeper}, GCN is essentially a special form of Laplacian smoothing. Therefore, a deeper GCN may bring potential concerns of over-smoothing, \emph{i.e.} the output features may be over-smoothed and the vertices from different clusters may become indistinguishable. In summary, the utilization of unlabeled information beyond the $K$-hop neighborhood remains an open problem. 

In this paper, we propose a new architecture that can discover much more information within unlabeled vertices and learn from the global graph topology. A key innovation is to marry Mean Teacher framework~\cite{tarvainen2017mean} into the classic GCN. Instead of merely supervising the propagated features in labeled nodes, we directly give chances to unlabeled nodes to ``speak up for themselves''. Specifically, SEGCN contains a student model and a teacher model. As a student, it not only learns to correctly predict the labeled nodes, but also tries to be consistent with teacher output on unlabeled nodes in more challenging settings, such as high dropout rates and graph collapse. As a teacher, it updates itself by averaging the student model weights. Since the teacher model operates under better settings such as low dropout rates and lossless graph, it generates more accurate predictions on both labeled and unlabeled nodes, thus being able to lead the student to learn in the next epoch. In such a mutual-promoting process, both labeled and unlabeled samples can be fully utilized for back-propagating effective gradient to train GCN.

To the best of our knowledge, this is the first time to introduce Mean Teacher strategy in the GCN design. Precisely, the main contributions of this work are summarized below. 
\begin{itemize}
\item By proposing to combine Mean Teacher with classic GCN, we emphasize the importance of exploitation of unlabeled nodes in graph-structured data classification.
\item Analogy to the noise added to student model in regular data, we successfully adapt Mean Teacher to graph-based data by designing new perturbation strategies for student model.
\item Our results are on par with the state-of-the-art methods on three node classification benchmarks in terms of accuracy, \emph{i.e.} Citeseer ($69.9\% \rightarrow 73.4\%$), Core ($80.4\% \rightarrow 83.5\%$) and Pubmed ($78.6\% \rightarrow 78.9\%$).
\end{itemize}

The rest of this paper is organized as follows. Section 2 discusses related work for GCN and Mean Teacher that provide the foundation for this paper. Then we propose the SEGCN model in Section 3. Section 4 presents an experimental study in which we compare our method with baseline and state-of-the-art results of benchmark datasets. Finally, we conclude with our contributions in Section 5.

\section{Preliminaries and Related Work}
We first provide a brief introduction to the required background. Then we review SSL with GCN and Mean Teacher, which provide fundamental theories for this paper.
\subsection{Spectral Graph Convolution}
There are two means to define convolution on graph, either from a spatial approach or from a spectral approach. This paper focuses on the latter. Based on the theory of Chung \emph{et al.}~\cite{chung1997spectral}, spectral GCNNs construct the convolution kernel on spectrum domain. They represent both the filter and the signal with the Fourier basis and multiply them, then transforms the result back into the discrete domain. However this model requires explicitly computing the eigenvectors of Laplacian matrix $L$, which is impractical for real large graphs. To circumvent this problem, it was suggested in~\cite{hammond2009wavelets} that approximate the spectral filter $g_{\theta'}$ with Chebyshev polynomials up to $Kth$ order:
\begin{equation}
  g_{\theta'} \star x \approx  \sum_{k=0}^{K} \theta_k' T_k(\tilde{L}) x \, ,
\label{eq:fourier-conv-approx}
\end{equation}
where $\theta^\prime \in \mathcal{R}^K$ is a vector of Chebyshev coefficients and $T_k(.)$ denotes the $Kth$ item of Chebyshev polynomials. $\tilde{L} = \frac{2}{\lambda_{\text{max}}}L-I_N$, where $I_N$ is a $N$ order diagonal matrix and $\lambda_{max}$ denotes the largest eigenvalue of $L$.

Recent proposed GCN~\cite{kipf2016semi} further simplifies this model by limiting K = 1 and approximating $\lambda_{max} \approx 2$. Given the adjacent matrix $A$ and the input feature $X$, the output of a single convolutional layer $Z$ can be represented as
\begin{equation}
  Z = \tilde{D}^{-\frac{1}{2}}\tilde{A}\tilde{D}^{-\frac{1}{2}}X\Theta \, ,
\label{eq:gcn-layer}
\end{equation}
where $\tilde{A} = A + I_N$, $\tilde{D}_{ii} = \sum_j \tilde{A}_{ij}$ and $\Theta$ denotes the trainable model parameters. Specifically, a two-layer GCN model can be defined as
\begin{equation}
Z= f(X,A) =  \mathrm{softmax}\!\left(\hat{A} \,\, \mathrm{ReLU}\!\left(\hat{A} X \Theta^{(0)} \right) \Theta^{(1)} \right) \, .
\label{eq:two-layer-gcn}
\end{equation}
where $\hat{A} = \tilde{D}^{-\frac{1}{2}}\tilde{A}\tilde{D}^{-\frac{1}{2}}$. This two-layer model forms the backbone of our SEGCN.

\subsection{Semi-supervised Learning with GCNs}
The above GCN model in Eq.~\ref{eq:two-layer-gcn} can be expediently used for SSL. However, as the analysis in Sec.~\ref{sec:introduction}, this method is limited to its small receptive field within few-hops neighborhood. Several recent methods are proposed to overcome such limitation~\cite{weston2012deep,abu2018N-GCN,verma2018feastnet,monti2017Monet}. 
Among these attempts, Random Walk is proven to be very effective to discover more remote cues~\cite{grover2016node2vec,perozzi2014deepwalk}. Moreover, attention mechanisms are introduced to emphasize the useful unlabeled information~\cite{Kiran2018attention,veličković2018GAT,shang2018edge}. To enable SSL on an extreme large graph, Liao \emph{et al.}~\cite{liao2018GPNN} extendse GCN by graph partitions.

\subsection{Semi-supervised Learning with Mean Teacher}
Mean Teacher~\cite{tarvainen2017mean} is one of the self-ensembling methods. The idea of a teacher model training a student is related to model compression~\cite{buciluǎ2006model} and distillation~\cite{hinton2015distilling}. Apart from other variants~\cite{bachman2014learning,laine2016temporal}, Mean Teacher averages model weights instead
of predictions and achieves excellent results in SSL. Other lines of work in self-ensembling focus on designing effective perturbation, including~\cite{gastaldi2017shake,huang2016stochasticdepth,wan2013dropconnect}.

\section{Proposed Method}
%This section presents our approach. We first introduce a framework organized by generations (Section 3.1). Then, we provide empirical analysis on why this framework trains deep networks better (Section 3.2), based on which we propose to set tolerant teachers to educate better students (Section 3.3).

\subsection{Combining GCN with Mean Teacher}
Graph Convolutional Network and Mean Teacher model are individually powerful. However, as we present in early sections, the former explores the unlabeled information by halves while the latter has only shown its ability on Euclidean data. In this section, we propose a new framework called SEGCN that combines the merits of both models while overcomes their limitations. Particularly, SEGCN contains a student model $f(\Theta_s)$ and a teacher model $f(\Theta_t)$, where $\Theta_s$ and $\Theta_t$ are the weights of the respective models. Given the labeled data $\D_L = \{(x^L_i, y^L_i)\}_{i=1}^{N_L}$ and unlabeled data $\D_U = \{x^U_i\}_{i=1}^{N_U}$, we first construct a normalized adjacent matrix $A$ upon these data according to their pairwise relations. Specific to our article classification task, the pairwise relations consist in a citation from one article to another. 

We formulate the overall loss of SEGCN from two aspects. On the one hand, the student should learn to minimize the cross-entropy loss under the supervision of labeled data in a noise-free environment. 
\newline
\begin{equation}
\label{eq:ce_loss}
\ell_{\text{CE}}(\Theta_s, A, x, y) = -\sum_{c=1}^C y_{c} \log f(A, x; \Theta_s)_c  \,,
\end{equation}
\newline
where $x$ denotes a labeled sample. The $f(A, x; \Theta_s)_c$ denotes the predicted probability from the student classifier on the class $c$. The $y_c$ denotes the ground truth probability of the class $c$.

On the other hand, the student classifier should be consistent with teacher's predictions when operates under small perturbations. In classic Mean Teacher on Euclidean-based data, the perturbation is usually added to original inputs $x$ to construct student inputs. Differently, the input fed in SEGCN are the \emph{Bag-of-Words} (\emph{BoW}) features distilled from articles. As a result, the traditional data augmentation strategies such as scaling, inversion or distortion are not available in our task. Instead, we innovatively construct such perturbation on both the graph structure and the model of student network. For the perturbation on graph structure $A$, we generate a ``collapsed graph'' by $A^\prime = P_A(A)$ where $P_A(.)$ is our perturbation function on the normalized adjacent matrix $A$. Similarly, we add noise to the original student model $f(.)$ by $f^\prime(.) = P_f(f(.))$ where $P_f$ is the perturbation function on model. These two perturbation functions will be detailed discussed in Sec.~\ref{sec:perturbation}. Given the \emph{disturbed} setting $\{A^\prime, f^\prime(.)\}$ for student and the \emph{original} setting $\{A, f(.)\}$ for teacher, the unsupervised consistency loss would penalizes the difference between the student's predicted probabilities $f^\prime(A^\prime, x; \Theta_s)$ and the teacher's $f(A, x; \Theta_t)$. In our paper, we formulate this loss as KL divergence.

\begin{figure}
\centering
\includegraphics[width=0.9\linewidth]{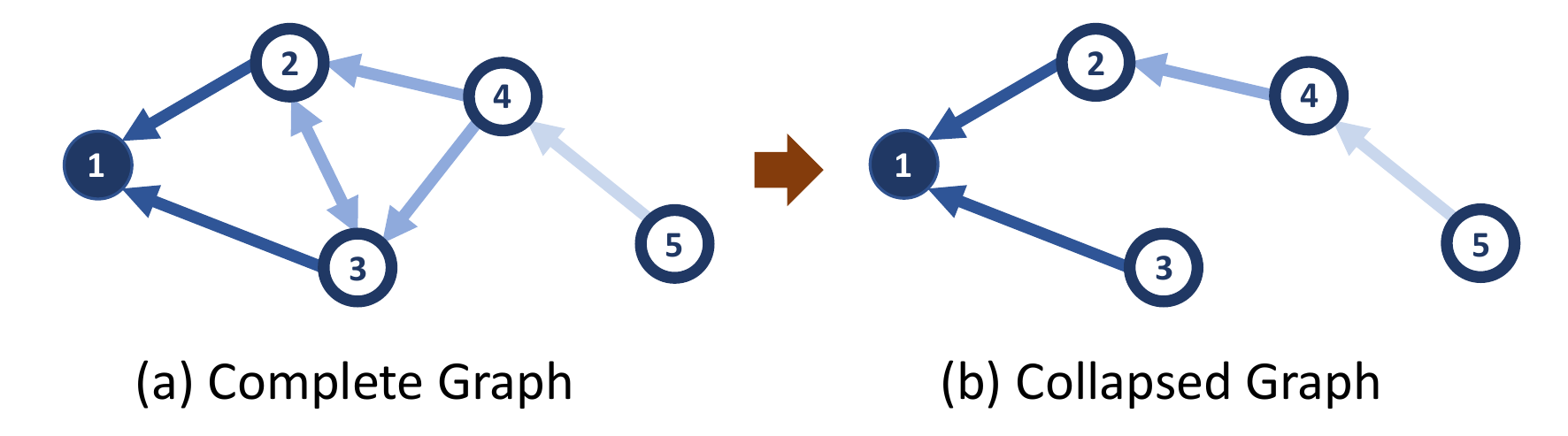}
\caption{Perturbation on graph structure. The arrows indicate the feature aggregation directions. A deeper blue arrow represents a larger kernel weight. \textbf{(a)} In original complete graph, the propagation paths between vertices are abundant, \emph{e.g.,} the vertex $5$ can propagate its feature to the labeled vertex $1$ via four different routes. \textbf{(b)} On the premise of no isolated node appears, $P_A(.)$ randomly cuts off the redundant edges in order to partially block the information transmission between vertices. Our perturbation strategy enables the student to generate worse predictions while keeping the intrinsic \emph{BoW} features unchanged.}
\label{fig:perturbation}
\end{figure}

\begin{equation}
\label{eq:cons_loss}
\ell_{\text{cons}}(\Theta_t, \Theta_s, A, A^\prime, x) = \text{KL}( f(A, x; \Theta_t) || f^\prime(A^\prime, x; \Theta_s) ) \,.
\end{equation}
\newline
With the above loss terms in Eq.~\ref{eq:ce_loss} and Eq.~\ref{eq:cons_loss}, the overall loss function of our approach can be written as
\newline
\begin{equation}
  \label{eq:overall_loss}
  \mathcal{L}(\Theta_t, \Theta_s, A, A^\prime, x, y) = 
  \underbrace{\sum_{(x, y) \in \D_L} \ell_{\text{CE}}}_{\mathcal{L}_{\text{Sup}}} +
  \lambda \underbrace{ \sum_{x \in \D_L \cup \D_U} \ell_{\text{cons}}}_{\mathcal{L}_{\text{Unsup}}},
\end{equation}
\newline
where the parameter $\lambda > 0$ controls the relative importance of the consistency term in the overall loss.

\begin{figure*}[ht]
\centering
\includegraphics[width=0.9\linewidth]{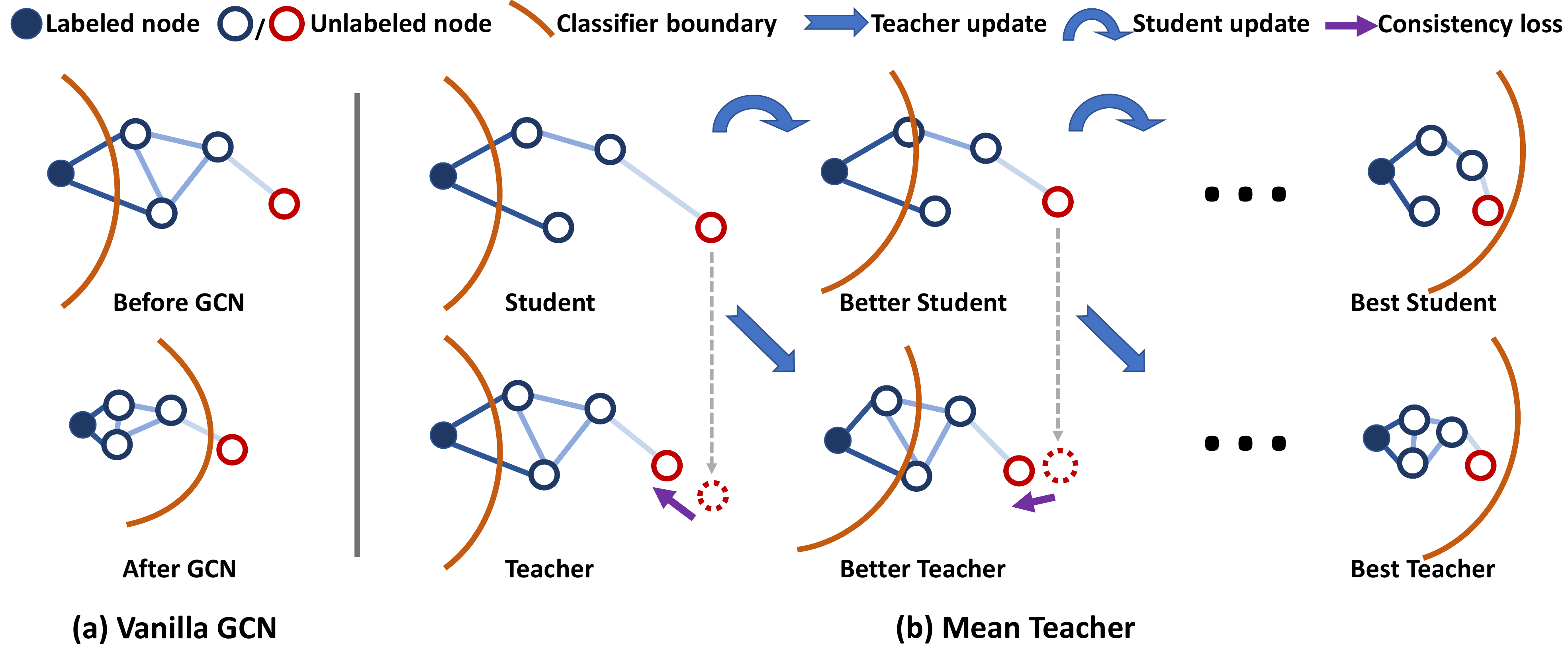}
\caption{(Best viewed in color.) \textbf{(a)} Vanilla GCN pushes decision boundaries away from the labeled sample. Since the labeled feature is mainly aggregated from adjacent vertices, the near samples can be well classified. However, a remote unlabeled node (red circle) is prone to be ignored by assigning a very small GCN kernel weight. Consequently, the vanilla model may be under-constrained and cannot be well adapted to these distant samples. \textbf{(b)} SEGCN pays more attentions on the unlabeled nodes than vanilla GCN. Via a mutual-promoting process leading by the consistency losses (\emph{e.g.,} the inconsistency between the solid/dotted red circles), SEGCN gives the chances to the unlabeled nodes as well to backpropagate effective gradient to train the model. Once the mutual-promoting process converges, even a remote unlabeled node can be well classified.}
\label{fig:phenomenon}
\end{figure*}

SEGCN is trained end-to-end to minimize the overall loss. In such scheme, the student can not only learn to distill knowledge from the labeled data under the supervised loss, but also pay much attention to the unlabeled vertices in order to come after the teacher. On the other side, through averaging the latest parameters in the student, the teacher is able to evolve itself since the gradient descent direction leading by consistency loss also applies to the update for the teacher model. As shown in Fig.~\ref{fig:phenomenon}, such a joint evolution between the dual GCNs paves the way for more thoroughly exploration on unlabeled information, which is otherwise not possible if solved alone in their traditional frameworks. 

\subsection{Perturbation in Student Model} \label{sec:perturbation}
This subsection aims to design effective perturbations on graph-based data. We do not follow the traditional ways of adding noise to raw inputs~\cite{french2017self}, which is popularly used on images and videos. The reason is two-fold. Firstly, the traditional data augmentation strategies on Euclidean data such as scaling, inversion or distortion are not suitable in article classification task. Secondly, some keywords are closely related to the article category. If we directly modify the \emph{BoW} feature vectors, the intrinsic cues for classifying an article may be lost. As a result, we instead propose two perturbation strategies on the graph structure (represented by adjacent matrix $A$) and GCN model respectively. 

The perturbation operation on graph structure, which is denoted as $P_A(.)$, is shown in Fig.\ref{fig:perturbation}. %On the premise of all vertices being connected, 
On the premise of no isolated node appears, $P_A(.)$ randomly cuts off the redundant edges in order to partially block the information transmission between vertices. The decreased number of connection makes the feature aggregation from neighbor vertices becoming even harder, thus enabling the student to generate inconsistent predictions with the teacher.  

Another function $P_f(.)$ denotes the perturbation on model itself. We implement $P_f(.)$ by appending a dropout layer~\cite{srivastava2014dropout} to each model. The dropout layer can drop different nodes in each time and obtaining two different output vectors from student and teacher. To force the student to operate in a harder environment, we give a higher dropout rate to the student than the teacher.  
In a nutshell, the two specifically designed perturbations enable the student to generate inconsistent predictions while keeping the intrinsic \emph{BoW} features unchanged. 

\subsection{Self-training in Mean Teacher}
Self-training is an effective scheme where the model is bootstrapped with additional labeled data obtained from its own highly confident predictions. This process is repeated until some termination conditions are reached. Although these methods are heuristic and have achieved much progress in semi-supervised learning, seeking for an appropriate threshold to select those high confident predictions is by no means easy. A strict threshold may reject most of the right predictions, thus leading to a degenerated self-training scheme. On the contrary, a loose threshold may bring about a poorer classifier since the wrongly imported pseudo labels can reinforce poor predictions. Traditionally, this threshold is set empirically and any prediction with a softmax score larger than the threshold will be selected as an extended labeled sample.

Different from the traditional methods which generate predictions merely from a single classifier, SEGCN is born with the dual classifiers and able to make predictions from two views, \emph{i.e.} the student view and the teacher view. This property motivates us to combine the different views aiming to select more robust pseudo labels. Particularly, we regard a prediction as a high confident result only if \emph{both} student and teacher give larger softmax scores than the threshold $t$ on the same class. Otherwise, if the two classifiers give inconsistent predictions, it indicates a probably incorrect prediction which will be excluded from the additional labeled data in this epoch. In our experiment, we verify that combining teacher and student can achieve a better self-training performance compared with the single model-based variant.

\begin{algorithm}[t]
    \caption{GCN in Mean Teacher framework}\label{alg:Mean Teacher}
    \label{alg:train SEGCN}
    \begin{algorithmic}[1]
    \STATE \textbf{Input :} \\
    \quad Two-layer GCN model $f(.)$ \\  
    \quad Normalized adjacent matrix $A$ \\  		
    \quad \emph{BoW} features and labels $(X^L, Y^L)$ \& $X^U$
    \STATE \textbf{Initialization :} \\
    \quad High-noise model $f^\prime(.) = P_f(f(.))$ \\
    \quad Collapsed graph $A^\prime = P_A(A)$ \\
    \quad Model weights $\Theta_t = \Theta_s$ from scratch \\
    \STATE \textbf{Repeat :}
    \STATE \quad $\mathcal{L}_{Sup} \leftarrow Eq.~\ref{eq:ce_loss}$  \\
    \STATE \quad $\mathcal{L}_{Unsup} \leftarrow Eq.~\ref{eq:cons_loss}$  \\
    \STATE \quad $\mathcal{L}_{Overall} \leftarrow Eq.~\ref{eq:overall_loss}$  \\
    \STATE \quad $\Theta_s \leftarrow ADAM$  \\
    \STATE \quad $\Theta_t \leftarrow Eq.~\ref{eq:moving_average}$  \\
    \STATE \quad Add high confident pseudo labels to $(X^L, Y^L)$\\
    \STATE \textbf{Until} $\mathcal{L}_{Overall}$ converges.
    \end{algorithmic}
\end{algorithm}

\subsection{Training SEGCN} \label{sec:training}
The training of SEGCN framework is essentially a mutual-promoting process. We detail the training step in Algorithm~\ref{alg:train SEGCN}. The student weight $\Theta_s$ is initialized from scratch while the teacher weight $\Theta_t$ is initialized by copying from $\Theta_s$. Given the labeled \emph{BoW} features $(X^L, Y^L)$ and the unlabeled ones $X^U$, the student is firstly trained to minimize the supervised cross-entropy loss on $X^L$ using a \emph{complete graph} $A$ and a \emph{low dropout model} $f(.)$. Then the student is forced to be consistent with the teacher predictions on all vertices $X^L \bigcup X^U$ using a \emph{collapsed graph} $A^\prime$ and a \emph{high dropout model} $f^\prime(.)$. Finally, the teacher updates its own model weights from the latest student model. In each iteration, the student network is optimized using ADAM~\cite{kingma2014adam}, while the weights of the teacher network are updated with an exponential moving average of those of the latest student, which is formulated as Eq.~\ref{eq:moving_average}.
\newline
\begin{equation}
  \label{eq:moving_average}
  \Theta_t^{e+1} = \alpha \Theta_t^{e} + (1-\alpha) \Theta_s^{e+1} \,,
\end{equation}
\newline
where $\alpha$ is a smoothing coefficient hyperparameter and $e$ denotes the current epoch.

Since the student and the teacher are both inaccurate in early epochs, the mutual learning between each other could lead to unstable predictions, which may deviate from our original intention. Therefore, traditional solution utilizes a two-stage training scheme. Namely, in first stage the student is merely trained on labeled data until it converges. Then the mutual learning is started up in the second stage. Differently, to enable an end-to-end scheme, we approximate the classic two-stage training process using a hyperparameter trick. We initialize small $\lambda$ and $\alpha$ in early epochs and progressively increase them during the training. In such one-stage scheme we can avoid the unwanted knowledge transmission between the immature partners in early epochs.

\section{Experiment} \label{sec:experiment}
\subsection{Datasets}
We experiment on three public available citation graph datasets: \textbf{Citeseer}, \textbf{Cora} and \textbf{Pubmed}. Table~\ref{tab:dataset} summarizes dataset statistics. A citation graph dataset consists of documents as nodes and citation links as directed edges. Each node has a human annotated topic from a finite set of classes and a feature vector. For Citeseer and Cora, the feature vector has binary entries indicating the presence/absence of the corresponding word from a dictionary. For the Pubmed dataset, the feature vector has real-values entries indicating Term Frequency-Inverse Document Frequency \emph{(TF-IDF)} of the corresponding word from a dictionary. Although the networks are directed, we use undirected versions of the graphs for all experiments, which is common in all baseline approaches.

Besides the three benchmarks above, we construct a new toy dataset called \textbf{Node5} in order to clearly showcase the effect of SEGCN on far-away nodes. It derives from a subset of Cora. For each class in Cora, we select 5 (1 labeled and 4 unlabeled) samples and link them as a complete graph structure presented in Fig.~\ref{fig:perturbation}. The experiment on this toy dataset aims to verify the phenomenon we present in Fig.~\ref{fig:phenomenon}.
\begin{table}[t]
\caption{Dataset statistics}
\label{tab:dataset}
\vspace{0.1cm}
\centering
\resizebox{\linewidth}{!}{
\begin{tabular}{lcccc}
\hline
\textbf{Dataset}    &    \textbf{\# Nodes}    &    \textbf{\# Edges}    &    \textbf{\# Classes}    &    \textbf{\# Feature Dim.} \\
\hline
Node5		&	35		&	42		&	7	&	1433	\\
Citeseer	&	3327	&	4732	&	6	&	3703    \\
Cora		&	2708	&	5429	&	7	&	1433    \\
Pubmed		&	19717	&	44338	&	3	&	500    \\
\hline
	
\end{tabular}
}
\end{table}

\subsection{Experimental Setup}
We use PyTorch for implementation. we train both baseline and SEGCN as two-layer networks described in GCN~\cite{kipf2016semi}, where the first layer outputs 16 dimensions per node and the second layer outputs the number of classes. For the student model, we use ADAM optimizer~\cite{kingma2014adam} with $\beta1 = 0.9$, $\beta2 = 0.999$, $weight decay = 0.0005$ and $dropout rate = 0.5$. While for the teacher, no dropout is applied. We fix the learning rate to 0.01 and train SEGCN for 1,000 epochs. As mentioned in Sec.~\ref{sec:training}, hyperparameter $\lambda$ in Eq.~\ref{eq:overall_loss} gradually increases from $0$ to $2$ while $\alpha$ in Eq.~\ref{eq:moving_average} increases from $0$ to $0.999$ in our best model. To import more pseudo label candidates when our model is stable, we gradually decrease the self-training threshold $t$ from $0.9$ to $0.7$. In all experiments, we utilize 500 samples as a validation set and evaluate prediction accuracy on a test set of 1,000 examples. These validation samples are used for capturing the model parameters at peak validation accuracy to avoid overfitting.

\subsection{Comparative Study}
\textbf{Fixed splits.} In the first experiment, we use the fixed data splits from the work of Yang \emph{et al.}~\cite{yang2016revisiting} as it is the standard benchmark data splits in literatures. Specifically, these experiments are run on the same fixed split of 20 labeled nodes for each class.  We present the classification accuracy on the mentioned three benchmarks in Table~\ref{tab:main} with comparisons to our baseline as well as the state-of-the-art semi-supervised classification methods. Except our own implemented baseline GCN model, the accuracy of the other comparative methods are all taken from existing literature.

\begin{table*}[ht]
\caption{Accuracy comparison with the state-of-art-methods under the setting of \textbf{Fixed / Random} data splits. $^*$ indicates our own implemented baseline.}
\label{tab:main}
\vspace{0.1cm}
\centering
%\resizebox{\linewidth}{!}{
\begin{tabular}{l|ccc|ccc}
\hline
& \multicolumn{3}{c}{\textbf{Fixed splits}} & \multicolumn{3}{|c}{\textbf{Random splits}}\\
\textbf{Method}  & \textbf{Citeseer} & \textbf{Cora} & \textbf{Pubmed} & \textbf{Citeseer} & \textbf{Cora} & \textbf{Pubmed}\\ 
\hline
%ManiReg & $60.1$ & $59.5$ & $70.7$ & - & - & - \\

%SemiEmb \small{(Weston et al. 2012)} & $59.6$ & $59.0$ & $71.1$ & - & - & - \\

%LP & $45.3$ & $68.0$ & $63.0$ & - & - & - \\

DeepWalk \small{(Perozzi et al. 2014)} & $43.2$ & $67.2$ & $65.3$ & $47.2$ & $70.2$ & $72.0$\\

node2vec \small{(Grover et al. 2016)} & $54.7$ & $74.9$ & $75.3$ & $47.3$ & $72.9$ & $72.4$\\

%ICA & $69.1$ & $75.1$ & $73.9$  & - & - & - \\

DCNN \small{(Atwood et al. 2016)} & - & $76.8$ & $73.0$  & - & - & - \\

Planetoid \small{(Yang et al. 2016)} & $64.7$ & $75.7$ & $77.2$ & - & - & - \\

GCN \small{(Kipf et al. 2016)} & $70.3$ & $81.5$ & $79.0$ & $67.9 \pm 0.5$ & $80.1 \pm 0.5$ & $\mathbf{78.9} \pm 0.7$\\

Graph-CNN \small{(Such et al. 2017)} & - & $76.3$ & -  & - & - & - \\

MoNet \small{(Monti et al. 2017)}  & - & $81.7 \pm 0.5$ & $78.8 \pm 0.3$  & - & - & - \\

Bootstrap \small{(Buchnik et al. 2017)} & $53.6$ & $78.4$ & $78.8$ & $50.3$ & $78.2$ & $75.6$\\

FeaStNet \small{(Verma et al. 2018)} & - & $81.6$ & $79.0$ & - & - & - \\

GPNN \small{(Liao et al. 2018)} & $69.7$ & $81.8$ & $79.3$  & $68.6 \pm 1.7$ & $79.9 \pm 2.4$ & $76.1 \pm 2.0$ \\

N-GCN \small{(Abu-El-Haija et al. 2018)} & $71.0$ & $81.8$ & $\mathbf{79.4}$ & - & - & - \\

F-GCN \small{(Vijayan et al. 2018)} & 72.3 & 79.0 & - & - & - & - \\

GAT \small{(Velikovi et al. 2018)} & $ 72.5 \pm 0.7$  & $83.0 \pm 0.7$  & $79.0 \pm 0.3$  & - & - & - \\

\hline

GCN$^*$ & $69.9$ & $80.4$ & $78.6$ & $66.8 \pm 0.7$ & $79.6 \pm 0.6$ & $78.3 \pm 0.7 $ \\

SEGCN & $\mathbf{73.4} \pm 0.7$ & $\mathbf{83.5} \pm 0.4$ & $78.9 \pm 0.7$ & $\mathbf{69.0} \pm 0.9$ & $\mathbf{80.8} \pm 1.0$ & $78.0 \pm 1.4 $\\

\hline
\end{tabular}
%}
\end{table*}

On the one hand, we observe that SEGCN can significantly outperforms baseline GCN method, which brings $3.5\%$, $3.1\%$ and $0.3\%$ improvement on Citeseer, Cora and Pubmed respectively. It implies that Mean Teacher can actually boost the classifier training. On the other hand, we compare SEGCN with the state-of-the-art methods including Deepwalk~\cite{perozzi2014deepwalk}, node2vec~\cite{grover2016node2vec}, DCNN~\cite{atwood2016DCNN}, Planetoid~\cite{yang2016revisiting}, Monet~\cite{monti2017Monet}, Bootstrap~\cite{buchnik2017bootstrapped}, Graph-CNN~\cite{such2017GraphCNN}, FeaStNet~\cite{verma2018feastnet}, GPNN~\cite{liao2018GPNN}, N-GCN~\cite{abu2018N-GCN},  F-GCN~\cite{vijayan2018F-GCN} and GAT~\cite{veličković2018GAT}. As it can be seen SEGCN performs best on Citeseer and Cora which yields new state-of-the-art accuracies, while falling short by only 0.5\% from the best on the Pubmed dataset. The great performance not only relies on the high-performance baseline GCN method, but also due to the proposed self-ensembling strategy applied in the training scheme.

\textbf{Random splits.} Next, following the setting of GCN~\cite{kipf2016semi} , we run experiments keeping the same size in labeled, validation, and test sets as in fixed splits, but now selecting those nodes uniformly at random. This, along with the fact that different topics have different numbers of nodes in it, means that the labels might not be spread evenly across the topics. For 20 such randomly drawn dataset splits, the average accuracy is shown in Table~\ref{tab:main} with the standard error. As we do not force an equal number of labeled data for each class, we observe that the performance degrades for all methods compared to fixed splits except Deepwalk~\cite{perozzi2014deepwalk}. Besides, SEGCN achieves best results among the state-of-the-art methods on Citeseer and Cora, yielding $69.0\%$ and $80.8\%$ respectively in accuracy. We also note that the variances of the accuracies become larger and the performance falls short on Pubmed than baseline. We will discuss this observation in next subsection.

\begin{figure*} 

\begin{minipage}[b]{.24\textwidth}
\includegraphics[height=3.2cm]{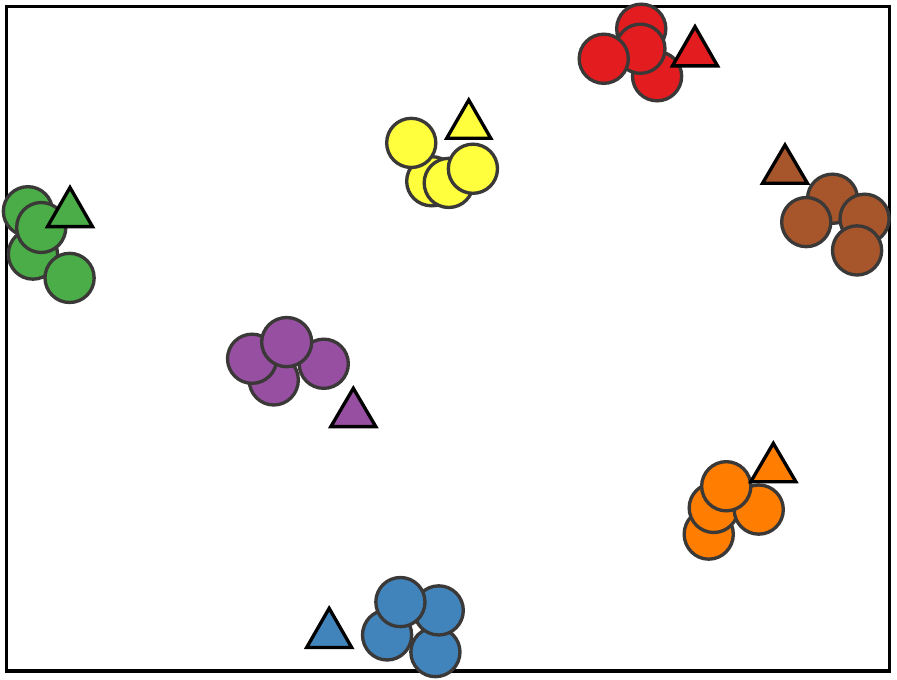}
\subcaption{}
\end{minipage}
\begin{minipage}[b]{.24\linewidth}
\includegraphics[height=3.2cm]{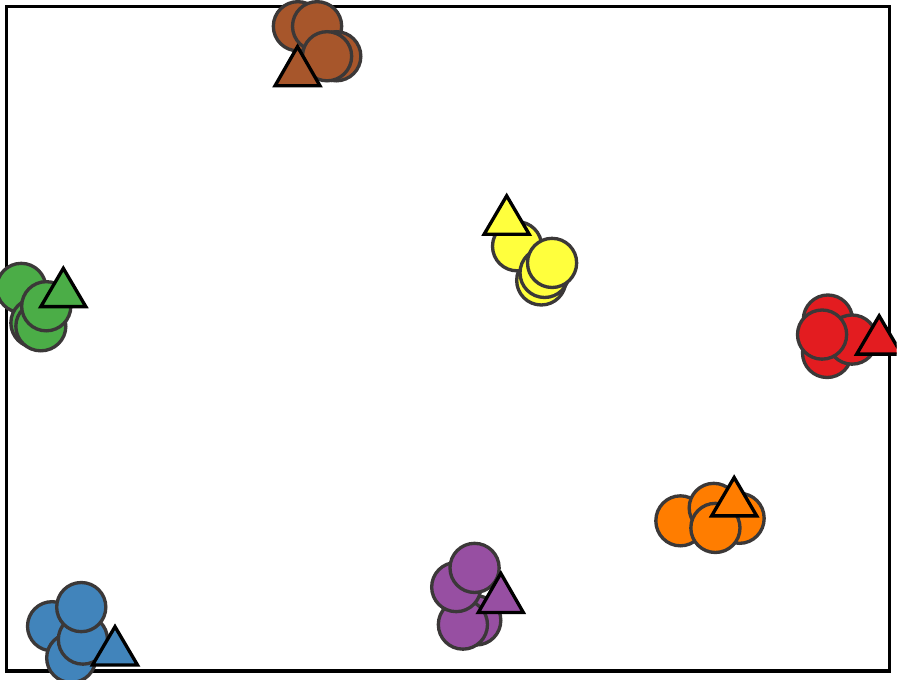}
\subcaption{}
\end{minipage}
\begin{minipage}[b]{.24\linewidth}
\includegraphics[height=3.2cm]{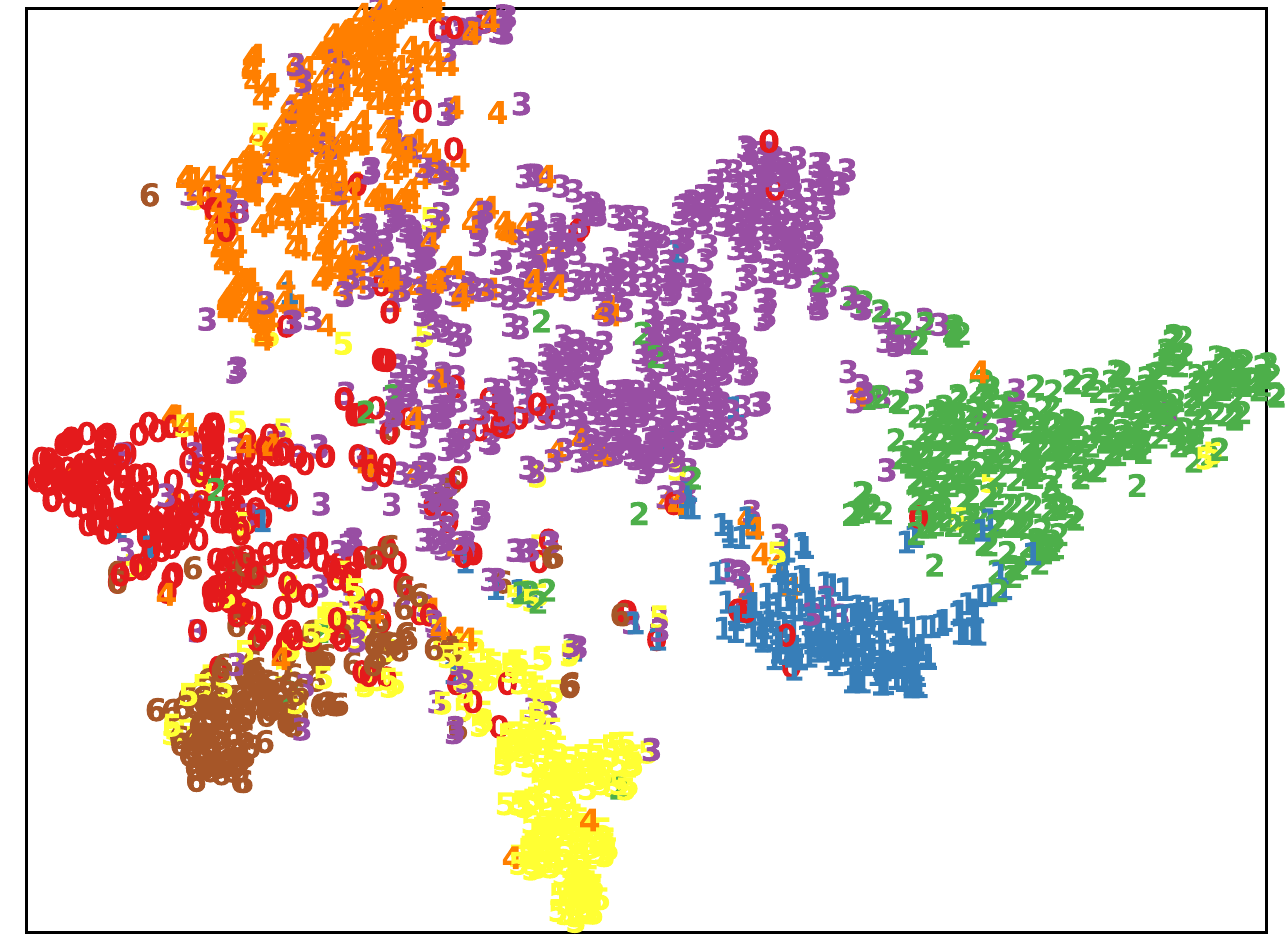}
\subcaption{}
\end{minipage}
\begin{minipage}[b]{.24\linewidth}
\includegraphics[height=3.2cm]{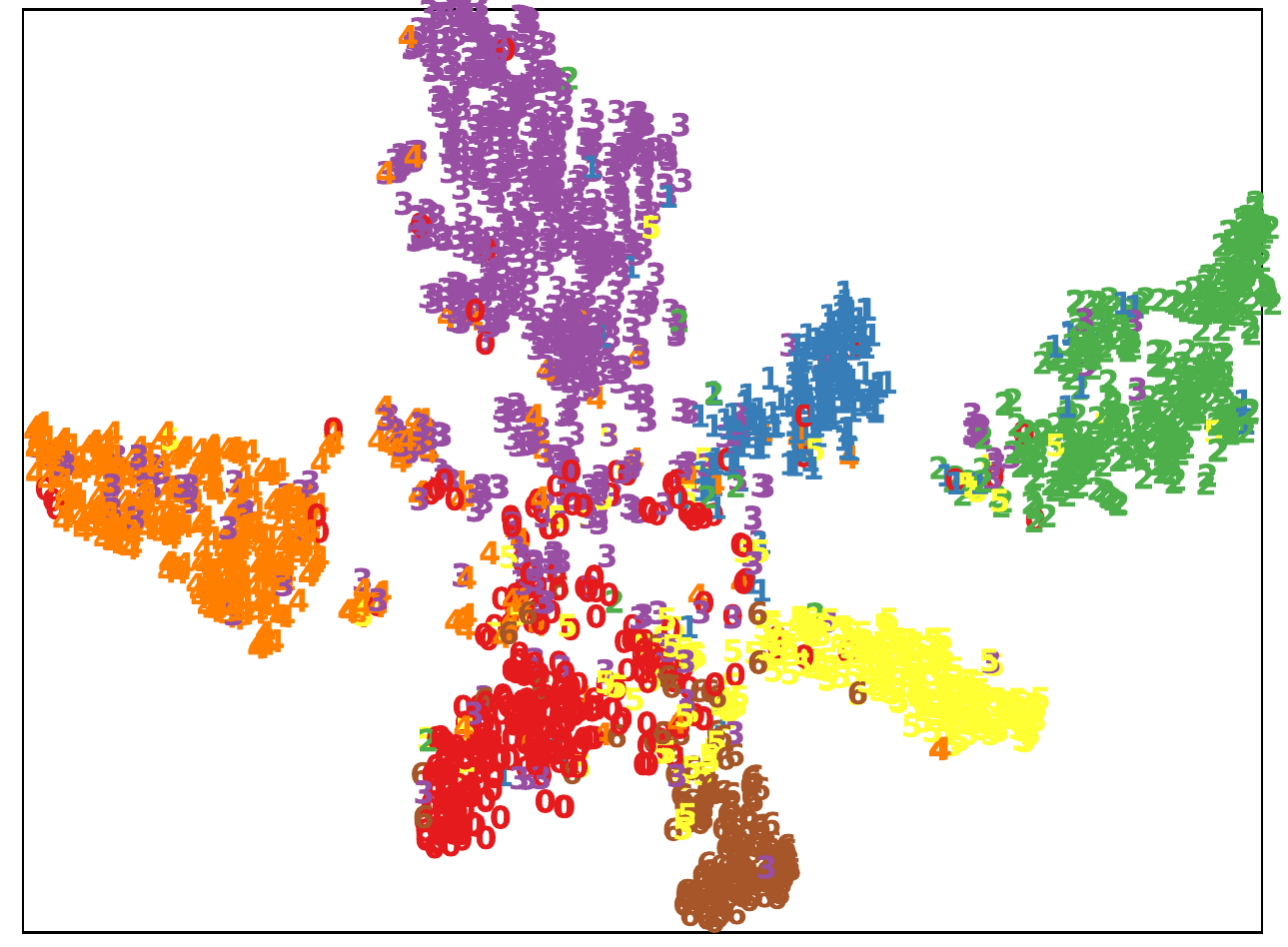}
\subcaption{}
\end{minipage}
\caption{Feature distribution analysis on Node5 ((a), (b)) and Cora ((c), (d)). We map the high-dimensional features outputted from the second layer to a 2-D space with t-SNE. (a)\&(c) are the result of vanilla GCN while (b)\&(d) are ours. Different color indicates different class. The triangles in (a)\&(b) denote the remote nodes (same as the red circles in Fig.~\ref{fig:phenomenon}).}
\label{fig:tSNE}
\end{figure*}

\textbf{More labeled samples on Pubmed.} We note that the improvement is relatively lower on Pubmed than that on Citeseer and Core. We suspect that the reason is due to the different label rates in the three benchmarks: the Pubmed dataset is relative large and the labeled rate is very low when we only select 20 training samples from each class. Consequently, the labeled nodes can only reach extremely limited scope in graph and can not give stable initial gradient directions to unlabeled vertices for the latter mutual-promoting process. Particularly, we observe that a few very low values appears under the random splits settings. These failure cases significantly decrease the average value and increase the variance. To clarify our hypothesis, we compare SEGCN with the state-of-the-art methods on Pubmed with more labeled samples over range \{50, 100, 200\}. As it can be seen SEGCN outperforms other methods in all cases with higher accuracies and lower variances. These results validate our hypothesis since a relative large labeled set can create stable initializations for mutual learning and self-training, which unlock the potential of SEGCN.

\subsection{Ablation Study}
To assess the importance of various aspects of the model, we run experiments on Cora under the setting of fixed splits, deactivating one or a few modules at a time while keeping the others activated. Table~\ref{tab:ablation} reports the classification accuracies under different ablations. To begin with, we test the two proposed perturbations $P_A(.)$ and $P_f(.)$ respectively, comparing with randomly erasing (\textbf{RE})~\cite{zhong2017random} the raw input features $X$. We observe that the direct perturbation on $X$ would hurt the final accuracy, dropping $0.3\%$ from the baseline and bringing about larger variance. While our proposed perturbations are helpful in improving the accuracy and $P_A$ is more effective than $P_f$. When combining the $P_A$ and $P_f$, SEGCN yields $82.4\%$ accuracy, which is higher than any single perturbation. It implies that $P_A$ and $P_f$ are complimentary and can lead the student to learn information from teacher. Then we test the self-training strategy with $P_A$ and $P_f$ fixed. We observe a significant improvement when combining student and teacher over a single model scheme, which implies that by using Mean Teacher we can produce more correct labeled sample candidates in self-training. Finally, SEGCN can yield new state-of-the-art accuracies when employing all the proposed modules. 

\begin{table}[t]
\caption{Accuracy comparison with the state-of-art-methods on Pubmed with varying \textbf{\# labeled sample} over range \{50, 100, 200\}. $^*$ indicates our own implemented baseline.}
\label{tab:pubmed}
\vspace{0.1cm}
\centering
\begin{tabular}{l|ccc}
\hline
& \multicolumn{3}{c}{\textbf{\# Labeled samples per class}}\\
\textbf{Method}  & \textbf{50} & \textbf{100} & \textbf{200}\\ \hline
DCNN	& -							& $82.6 \pm 0.3$ 			& - \\
N-GCN	& -							& $83.0 \pm 0.4$ 			& - \\
GCN$^*$	& $80.9 \pm 0.3$			& $81.8 \pm 0.3$ 			& $84.1 \pm 0.2$ \\
SEGCN	& $\mathbf{81.7} \pm 0.5$ 	& $\mathbf{83.8} \pm 0.4$ 	& $\mathbf{84.7} \pm 0.4$ \\
\hline
\end{tabular}
\end{table}

\subsection{Feature Distribution Analysis}
In this section, We aim to further prove the effectiveness of SEGCN via feature distribution analysis. To this end, we map the high-dimensional features distilled from the second layer into a 2-D space with t-SNE. Fig.~\ref{fig:tSNE} shows the t-SNE visualization, in which (a) and (b) represent the results of vanilla GCN and SEGCN respectively on Node5.  As depicted in (a) and (b), GCN maps the feature of the remote vertex far from the cluster center while SEGCN can  map it near. As a result, SEGCN significantly eases the classification task. The distribution distinction can be also observed on Cora shown in (c) and (d). These visualization results further validate the effectiveness of the mutual learning we described in Fig.~\ref{fig:phenomenon}. To sum up, SEGCN can capitalize on the every nodes' information to train a better classifier. It is effective for those remote unlabeled vertices, which are hard to be classified with vanilla GCN.

\section{Conclusion}
In this paper, we propose a self-ensembling framework called SEGCN for semi-supervised learning on graph-based data. Apart from the vanilla GCN, SEGCN can directly explore unlabeled information via the mutual learning between student and teacher, thus enabling every labeled and unlabeled nodes to backpropagate effective gradient to train the model. To the best of our knowledge, this is the first work that integrates the Mean Teacher model into GCN to boost the semi-supervised node classification. The extensive experiments on toy and public datasets show that SEGCN precedes the baseline model significantly and is on par with the state-of-the-art methods in tasks of article classification. 

\begin{table}[t]
\caption{Ablation study on Cora.}
\label{tab:ablation}
\vspace{0.095cm}
\centering
\begin{tabular}{ccc|cc|c}
\hline
\multicolumn{3}{c|}{\textbf{Perturbation}} & \multicolumn{2}{c|}{\textbf{Self-training}} & \\
$\mathbf{RE(.)}$ & $\mathbf{P_A(.)}$  & $\mathbf{P_f(.)}$ & \textbf{T} & \textbf{S\&T} & \textbf{Accuracy}\\ 
\hline
$\surd$	&			&			&			&			& $80.1 \pm 0.6$ \\
		& $\surd$	&			&			&			& $81.7 \pm 0.2$ \\
		&			& $\surd$	&			&			& $80.9 \pm 0.1$ \\
		&$\surd$	& $\surd$	&			&			& $82.4 \pm 0.2$ \\
		&$\surd$	& $\surd$	& $\surd$	&			& $83.1 \pm 0.6$ \\
		&$\surd$	& $\surd$	&			& $\surd$	& $\mathbf{83.5 \pm 0.4}$ \\
\hline
\end{tabular}
\end{table}

\bibliographystyle{aaai}
\bibliography{ref}

\begin{thebibliography}{}

\bibitem[\protect\citeauthoryear{Abu-El-Haija \bgroup et al\mbox.\egroup
  }{2018}]{abu2018N-GCN}
Abu-El-Haija, S.; Kapoor, A.; Perozzi, B.; and Lee, J.
\newblock 2018.
\newblock N-gcn: Multi-scale graph convolution for semi-supervised node
  classification.
\newblock {\em arXiv preprint arXiv:1802.08888}.

\bibitem[\protect\citeauthoryear{Atwood and Towsley}{2016}]{atwood2016DCNN}
Atwood, J., and Towsley, D.
\newblock 2016.
\newblock Diffusion-convolutional neural networks.
\newblock In {\em NIPS},  1993--2001.

\bibitem[\protect\citeauthoryear{Bachman, Alsharif, and
  Precup}{2014}]{bachman2014learning}
Bachman, P.; Alsharif, O.; and Precup, D.
\newblock 2014.
\newblock Learning with pseudo-ensembles.
\newblock In {\em Advances in Neural Information Processing Systems},
  3365--3373.

\bibitem[\protect\citeauthoryear{Buchnik and
  Cohen}{2017}]{buchnik2017bootstrapped}
Buchnik, E., and Cohen, E.
\newblock 2017.
\newblock Bootstrapped graph diffusions: Exposing the power of nonlinearity.
\newblock {\em arXiv preprint arXiv:1703.02618}.

\bibitem[\protect\citeauthoryear{Buciluǎ, Caruana, and
  Niculescu-Mizil}{2006}]{buciluǎ2006model}
Buciluǎ, C.; Caruana, R.; and Niculescu-Mizil, A.
\newblock 2006.
\newblock Model compression.
\newblock In {\em KDD},  535--541.

\bibitem[\protect\citeauthoryear{Caelles \bgroup et al\mbox.\egroup
  }{2017}]{caelles2017one}
Caelles, S.; Maninis, K.-K.; Pont-Tuset, J.; Leal-Taix{\'e}, L.; Cremers, D.;
  and Van~Gool, L.
\newblock 2017.
\newblock One-shot video object segmentation.
\newblock In {\em CVPR},  5320--5329.

\bibitem[\protect\citeauthoryear{Chung and Graham}{1997}]{chung1997spectral}
Chung, F.~R., and Graham, F.~C.
\newblock 1997.
\newblock {\em Spectral graph theory}.
\newblock Number~92. American Mathematical Soc.

\bibitem[\protect\citeauthoryear{Dai and Le}{2015}]{dai2015semi}
Dai, A.~M., and Le, Q.~V.
\newblock 2015.
\newblock Semi-supervised sequence learning.
\newblock In {\em NIPS},  3079--3087.

\bibitem[\protect\citeauthoryear{Defferrard, Bresson, and
  Vandergheynst}{2016}]{defferrard2016GCNN}
Defferrard, M.; Bresson, X.; and Vandergheynst, P.
\newblock 2016.
\newblock Convolutional neural networks on graphs with fast localized spectral
  filtering.
\newblock In {\em NIPS},  3844--3852.

\bibitem[\protect\citeauthoryear{French, Mackiewicz, and
  Fisher}{2017}]{french2017self}
French, G.; Mackiewicz, M.; and Fisher, M.
\newblock 2017.
\newblock Self-ensembling for visual domain adaptation.
\newblock {\em arXiv preprint arXiv:1706.05208}.

\bibitem[\protect\citeauthoryear{Gastaldi}{2017}]{gastaldi2017shake}
Gastaldi, X.
\newblock 2017.
\newblock Shake-shake regularization.
\newblock {\em arXiv preprint arXiv:1705.07485}.

\bibitem[\protect\citeauthoryear{Grover and
  Leskovec}{2016}]{grover2016node2vec}
Grover, A., and Leskovec, J.
\newblock 2016.
\newblock node2vec: Scalable feature learning for networks.
\newblock In {\em KDD},  855--864.

\bibitem[\protect\citeauthoryear{Hammond, Vandergheynst, and
  Gribonval}{2009}]{hammond2009wavelets}
Hammond, D.~K.; Vandergheynst, P.; and Gribonval, R.
\newblock 2009.
\newblock Wavelets on graphs via spectral graph theory.
\newblock {\em arXiv preprint arXiv:0912.3848}.

\bibitem[\protect\citeauthoryear{Hinton, Vinyals, and
  Dean}{2015}]{hinton2015distilling}
Hinton, G.; Vinyals, O.; and Dean, J.
\newblock 2015.
\newblock Distilling the knowledge in a neural network.
\newblock {\em arXiv preprint arXiv:1503.02531}.

\bibitem[\protect\citeauthoryear{Huang \bgroup et al\mbox.\egroup
  }{2016}]{huang2016stochasticdepth}
Huang, G.; Sun, Y.; Liu, Z.; Sedra, D.; and Weinberger, K.~Q.
\newblock 2016.
\newblock Deep networks with stochastic depth.
\newblock In {\em ECCV},  646--661.

\bibitem[\protect\citeauthoryear{Kingma and Ba}{2014}]{kingma2014adam}
Kingma, D.~P., and Ba, J.
\newblock 2014.
\newblock Adam: A method for stochastic optimization.
\newblock {\em arXiv preprint arXiv:1412.6980}.

\bibitem[\protect\citeauthoryear{Kipf and Welling}{2017}]{kipf2016semi}
Kipf, T.~N., and Welling, M.
\newblock 2017.
\newblock Semi-supervised classification with graph convolutional networks.
\newblock In {\em ICLR}.

\bibitem[\protect\citeauthoryear{Laine and Aila}{2016}]{laine2016temporal}
Laine, S., and Aila, T.
\newblock 2016.
\newblock Temporal ensembling for semi-supervised learning.
\newblock {\em arXiv preprint arXiv:1610.02242}.

\bibitem[\protect\citeauthoryear{Li \bgroup et al\mbox.\egroup
  }{2018}]{li2018adaptive}
Li, R.; Wang, S.; Zhu, F.; and Huang, J.
\newblock 2018.
\newblock Adaptive graph convolutional neural networks.
\newblock {\em arXiv preprint arXiv:1801.03226}.

\bibitem[\protect\citeauthoryear{Li, Han, and Wu}{2018}]{li2018deeper}
Li, Q.; Han, Z.; and Wu, X.-M.
\newblock 2018.
\newblock {Deeper Insights into Graph Convolutional Networks for
  Semi-Supervised Learning}.
\newblock In {\em AAAI}.

\bibitem[\protect\citeauthoryear{Liao \bgroup et al\mbox.\egroup
  }{2018}]{liao2018GPNN}
Liao, R.; Brockschmidt, M.; Tarlow, D.; Gaunt, A.~L.; Urtasun, R.; and Zemel,
  R.
\newblock 2018.
\newblock Graph partition neural networks for semi-supervised classification.
\newblock In {\em ICLR Workshop}.

\bibitem[\protect\citeauthoryear{Monti \bgroup et al\mbox.\egroup
  }{2017}]{monti2017Monet}
Monti, F.; Boscaini, D.; Masci, J.; Rodol{\`a}, E.; Svoboda, J.; and Bronstein,
  M.~M.
\newblock 2017.
\newblock Geometric deep learning on graphs and manifolds using mixture model
  cnns.
\newblock In {\em CVPR},  5425--5434.

\bibitem[\protect\citeauthoryear{Papandreou \bgroup et al\mbox.\egroup
  }{2015}]{papandreou2015weakly}
Papandreou, G.; Chen, L.-C.; Murphy, K.~P.; and Yuille, A.~L.
\newblock 2015.
\newblock Weakly-and semi-supervised learning of a deep convolutional network
  for semantic image segmentation.
\newblock In {\em ICCV},  1742--1750.

\bibitem[\protect\citeauthoryear{Perozzi, Al-Rfou, and
  Skiena}{2014}]{perozzi2014deepwalk}
Perozzi, B.; Al-Rfou, R.; and Skiena, S.
\newblock 2014.
\newblock Deepwalk: Online learning of social representations.
\newblock In {\em KDD},  701--710.

\bibitem[\protect\citeauthoryear{Rahimi, Cohn, and
  Baldwin}{2018}]{rahimi2018semi}
Rahimi, A.; Cohn, T.; and Baldwin, T.
\newblock 2018.
\newblock Semi-supervised user geolocation via graph convolutional networks.
\newblock {\em arXiv preprint arXiv:1804.08049}.

\bibitem[\protect\citeauthoryear{Shang \bgroup et al\mbox.\egroup
  }{2018}]{shang2018edge}
Shang, C.; Liu, Q.; Chen, K.-S.; Sun, J.; Lu, J.; Yi, J.; and Bi, J.
\newblock 2018.
\newblock Edge attention-based multi-relational graph convolutional networks.
\newblock {\em arXiv preprint arXiv:1802.04944}.

\bibitem[\protect\citeauthoryear{Srivastava \bgroup et al\mbox.\egroup
  }{2014}]{srivastava2014dropout}
Srivastava, N.; Hinton, G.; Krizhevsky, A.; Sutskever, I.; and Salakhutdinov,
  R.
\newblock 2014.
\newblock Dropout: a simple way to prevent neural networks from overfitting.
\newblock {\em The Journal of Machine Learning Research} 15(1):1929--1958.

\bibitem[\protect\citeauthoryear{Such \bgroup et al\mbox.\egroup
  }{2017}]{such2017GraphCNN}
Such, F.~P.; Sah, S.; Dominguez, M.~A.; Pillai, S.; Zhang, C.; Michael, A.;
  Cahill, N.~D.; and Ptucha, R.
\newblock 2017.
\newblock Robust spatial filtering with graph convolutional neural networks.
\newblock {\em IEEE Journal of Selected Topics in Signal Processing}
  11(6):884--896.

\bibitem[\protect\citeauthoryear{Tarvainen and
  Valpola}{2017}]{tarvainen2017mean}
Tarvainen, A., and Valpola, H.
\newblock 2017.
\newblock Mean teachers are better role models: Weight-averaged consistency
  targets improve semi-supervised deep learning results.
\newblock In {\em NIPS},  1195--1204.

\bibitem[\protect\citeauthoryear{Thekumparampil \bgroup et al\mbox.\egroup
  }{2018}]{Kiran2018attention}
Thekumparampil, K.~K.; Wang, C.; Oh, S.; and Li, L.-J.
\newblock 2018.
\newblock Attention-based graph neural network for semi-supervised learning.
\newblock {\em arXiv preprint arXiv:1803.03735}.

\bibitem[\protect\citeauthoryear{Veličković \bgroup et al\mbox.\egroup
  }{2018}]{veličković2018GAT}
Veličković, P.; Cucurull, G.; Casanova, A.; Romero, A.; Liò, P.; and Bengio,
  Y.
\newblock 2018.
\newblock Graph attention networks.
\newblock In {\em ICLR}.

\bibitem[\protect\citeauthoryear{Verma and Boyer}{2018}]{verma2018feastnet}
Verma, N., and Boyer, E.
\newblock 2018.
\newblock Feastnet: Feature-steered graph convolutions for 3d shape analysis.
\newblock In {\em CVPR}.

\bibitem[\protect\citeauthoryear{Vijayan \bgroup et al\mbox.\egroup
  }{2018}]{vijayan2018F-GCN}
Vijayan, P.; Chandak, Y.; Khapra, M.~M.; and Ravindran, B.
\newblock 2018.
\newblock Fusion graph convolutional networks.
\newblock {\em arXiv preprint arXiv:1805.12528}.

\bibitem[\protect\citeauthoryear{Wan \bgroup et al\mbox.\egroup
  }{2013}]{wan2013dropconnect}
Wan, L.; Zeiler, M.; Zhang, S.; Le~Cun, Y.; and Fergus, R.
\newblock 2013.
\newblock Regularization of neural networks using dropconnect.
\newblock In {\em ICML},  1058--1066.

\bibitem[\protect\citeauthoryear{Wang \bgroup et al\mbox.\egroup
  }{2018}]{wang2018dynamic}
Wang, Y.; Sun, Y.; Liu, Z.; Sarma, S.~E.; Bronstein, M.~M.; and Solomon, J.~M.
\newblock 2018.
\newblock Dynamic graph cnn for learning on point clouds.
\newblock {\em arXiv preprint arXiv:1801.07829}.

\bibitem[\protect\citeauthoryear{Weston \bgroup et al\mbox.\egroup
  }{2012}]{weston2012deep}
Weston, J.; Ratle, F.; Mobahi, H.; and Collobert, R.
\newblock 2012.
\newblock Deep learning via semi-supervised embedding.
\newblock In {\em Neural Networks: Tricks of the Trade}. Springer.
\newblock  639--655.

\bibitem[\protect\citeauthoryear{Yang, Cohen, and
  Salakhudinov}{2016}]{yang2016revisiting}
Yang, Z.; Cohen, W.; and Salakhudinov, R.
\newblock 2016.
\newblock Revisiting semi-supervised learning with graph embeddings.
\newblock In {\em ICML}.

\bibitem[\protect\citeauthoryear{Zhong \bgroup et al\mbox.\egroup
  }{2017}]{zhong2017random}
Zhong, Z.; Zheng, L.; Kang, G.; Li, S.; and Yang, Y.
\newblock 2017.
\newblock Random erasing data augmentation.
\newblock {\em arXiv preprint arXiv:1708.04896}.

\end{thebibliography}

\end{document}